# Contextual Pictograms for Repairing Multilingual ASR Errors

*Mariana Hernandez[1], Hector Luna[1,2], Carlos Bruno[3], Luis Ricardo Santos[3,4], Elena Simoes[1,2]*

[1] Department Signal Theory, Telematics and Communications, University of Granada
[2] Brain, Mind, and Behavior Research Center (CIMCYC), University of Granada
[3] Faculty of Psychology and Education Sciences, University of Coimbra
[4] Institute of Intelligent Systems and Robotics, CNRS, Sorbonne Université
lrsantos@fpce.uc.pt, esimoes@ugr.es

## Abstract

Speech transcription errors can be difficult to recognize when competing word hypotheses are unfamiliar or visually similar. We describe an interface that combines alternative transcripts, replay and contextually checked pictograms for selected uncertain concepts. The system will initially support French and Spanish functional utterances. A counterbalanced study will assess whether pictorial feedback helps listeners recover the speaker's intended action beyond text alternatives alone. We will also quantify semantic mismatch and additional interaction time, since an inappropriate image can reinforce the wrong interpretation rather than repair it.

## 1. Introduction

Visual symbols are often proposed as an intuitive supplement to text, yet they can be as ambiguous as the words they illustrate. Norming data for ARASAAC pictograms show relatively low transparency, that is, many symbols are hard to guess without their referent, with nouns faring best [1]; earlier work in augmentative and alternative communication (AAC) likewise found that the perceived link between a symbol and its referent varies across symbol systems and across parts of speech [2]. We ask a narrower question: when an ASR transcript contains an uncertain concrete word, does a context-checked pictogram help a listener select the intended alternative?

Automatic speech recognition has improved across languages, that it has even been used for other applications [3], [4], [5], [6], but a transcription can remain uncertain in noisy, accented or code-switched speech. Large-scale speech recognition systems offer a practical baseline for the task [7], yet their errors are not evenly spread: error rates have been shown to differ across speaker groups [8], code-switching remains a distinct difficulty [9], and large recognizers can hallucinate fluent phrases that were never spoken [10]. In communication settings, displaying the most probable string without showing uncertainty may lead users to accept an incorrect name, negation or instruction as if it were verified, an instance of the well-documented tendency to over-rely on automated output [11]. Marking uncertainty is not a complete answer either: in dictation, shading low-confidence words helped users find errors only when the confidence measure was accurate, and users tended to trust the shading even when it was wrong [12]. A repair aid therefore has to be judged by what it does when its own evidence is imperfect.

Visual repair could give speakers and recipients a second route to confirm intended meaning. Different conditions have been shown to rely on visual processing to compensate for other impairments [13], [14], [15], [16]. Prior multilingual work has mapped text concepts to pictograms and found that coverage and semantic suitability differ by language [17]. Related research on reading scaffolds indicates that additional symbols can help some users while increasing coordination demands for others [18]. A parallel line of work translates speech directly into pictogram sequences for people with language impairments, using ARASAAC pictograms and a French corpus of aligned speech, text and pictogram sequences [19], [20]. Those systems treat the pictogram sequence as the message itself. Our setting differs in two ways: the pictograms are an optional, local check on a few uncertain spans of a transcript that remains the primary channel, and the person judging the image is a listener who must choose between competing recognizer hypotheses rather than a reader of the whole sequence. The design problem is therefore deciding when a visual confirmation is likely to be informative.

## 2. System Concept

A speech recognizer will supply a transcript and alternatives for selected low-confidence spans, in the manner of the alternatives encoded in word lattices and confusion networks [21]. A repair interface can display the alternatives as text, play a local audio crop, or show a reviewed pictogram for a concrete concept. Multimodal correction has a long history in speech interfaces: switching to another modality was found to speed up the correction of dictation errors, and users came to favor the modality in which correction was most accurate [22]. That work concerned input modalities for re-entering text, whereas the added modality here is on the output side and serves disambiguation, so we test the benefit rather than assume that it transfers. The system will avoid mapping abstract words to misleading symbols merely to increase pictogram coverage. For the first version, the repair dialogue will operate in French and Spanish, with a later exploratory extension to Portuguese. Figure 1 summarizes the resulting repair flow.

Confirmation will focus on task-relevant spans: locations, objects, people and polarity-bearing terms. Anchoring repair in task content is consistent with how people recover when a recognizer corrupts a partner's speech: they tend to keep the dialogue moving on task content instead of explicitly signalling a misunderstanding [23]. A content word may be illustrated only if the mapping has passed review for that language and local context. The restriction to concrete concepts also follows the norming evidence: in the largest ARASAAC study to date, nouns had the highest transparency and translucency, and imageability and concreteness were the strongest correlates of both [1]. Polarity-bearing terms and temporal relations are abstract, so they will usually be confirmed through text and replay rather than an image. If no adequate image exists, the interface will show text and replay controls. The recipient or speaker can always dismiss an alternative without selecting an icon.

The pictogram lookup will use the candidate phrase rather than an isolated word as its context. For polysemous items, a reviewer will either approve one image for the full phrase or prohibit pictographic confirmation. We deliberately do not rely on automatic sense assignment here, since word sense disambiguation is a long-standing open problem whose errors would translate directly into wrong images [24]. This selective policy is expected to produce fewer icons than a coverage-maximizing baseline. A second risk concerns uneven illustration across competing candidates. Visual context is integrated rapidly during spoken-language comprehension [25], so an image attached to only one candidate may pull the choice toward it regardless of the audio. French sans 'without', sang 'blood' and cent 'hundred' are homophones, and of the three only sang is easy to depict as a concrete object. We will therefore log whether each candidate set was illustrated fully, partially or not at all and analyze partial sets as a distinct risk; if they prove harmful in the pilot, the interface can show a candidate set with images only when every member can be illustrated.

The repair policy will run after a fixed recognizer rather than modifying the acoustic model. We will retain word hypotheses and local timestamps so that uncertain spans can be replayed without repeating the full utterance. Alternative text candidates will be limited to a short list whose linguistic form matches the surrounding sentence. When the recognizer lacks calibrated confidence, we will use a held-out development set to estimate whether candidate rankings are suitable for deciding when to interrupt. This component-level check matters because a seemingly clear transcript can conceal a badly calibrated probability. Confidence estimation for speech recognition has an extensive literature [26], and we will treat the decision to interrupt as a reject option, in which the system defers to the person only when the top hypothesis is insufficiently reliable [27]. Because modern neural networks are often poorly calibrated [28], we will report per-language reliability diagrams and expected calibration error on the development set before fixing the interruption threshold. The stakes are practical: a confidence cue derived from an inaccurate measure can steer users away from real errors [12].

Visual alternatives will be evaluated separately for each language. A pictogram with the correct broad topic may still misrepresent number, role or action; it will not be accepted simply because it resembles a keyword. A reviewer will judge the target word in its sentence context, and the selected image will carry a pointer to the intended lexical sense. Research on illustrated text similarly finds that pictures aid learning when they are closely tied to the content rather than merely decorative [29]; the reviewer's sentence-level judgment is meant to enforce exactly that property. Where the repository lacks a suitable representation, the interface will abstain from showing an image. Such abstention is a designed safety behavior, not a missing result to be repaired by inventing a symbol. Coverage and abstention rates will be reported per language, because coverage and suitability are known to differ between languages [17].

# 3. Planned Evaluation

We will develop a small test set of everyday requests recorded by multiple speakers in French and Spanish. Background noise will be added at controlled levels to create uncertain ASR spans; to make the distortions reproducible, noise will be drawn from a public corpus such as MUSAN [30] and mixed at fixed signal-to-noise ratios. Each source message will be paired with a known intent representation, enabling evaluation of whether the corrected transcript preserves the requested action and its arguments. For the schema we can follow existing multilingual spoken-language-understanding resources: Speech-MASSIVE, for instance, provides read speech with intent and slot annotations in French, Spanish and European Portuguese [31]. Because its sentences are read rather than elicited requests, it is better suited as a source of development material than as the main test set, which will use our own recordings.

Participants will resolve uncertain messages under three displays: top-one transcript, transcript with alternatives, and transcript with alternatives plus validated pictograms. We will measure critical-word correction, task-intent recovery, time and unnecessary repair actions. A behavioral test suite will explicitly include negation, named entities and swapped temporal order, following the general approach of targeted linguistic capability testing [32]. Test items will be built as minimal pairs in which the competing hypotheses differ in one critical element, for example the French homophones sans, sang and cent, the Spanish pair pero 'but' and perro 'dog', or temporal contrasts such as avant 'before' and après 'after'. Such pairs also exercise the policy boundary, because some members can be illustrated and others cannot.

Mixed-effects analysis will account for speaker, listener and message as crossed random effects, with a random-effects structure as rich as the data support [33]. We will report outcomes separately by language and error category. The number of participants and the final noise levels will be refined in pilot testing rather than selected retrospectively to maximize differences between interfaces, a practice known to inflate false-positive rates [34]. Pilot data will also inform a simulation-based power analysis for mixed models with crossed random effects [31], and the analysis plan will be fixed before the main data collection.

We will count semantic mismatches separately from missing pictograms. A system that returns an image for nearly every word may score well on coverage while producing more incorrect confirmations. The main analysis will contrast useful repairs with visual-induced errors. Table 1 defines how each trial will be classified so that these two quantities, and the cost of unnecessary interventions, can be read from the same data.

**Table 1.** Classification of each repair trial. Every category is crossed with pictogram status (shown, withheld by the abstention rule, or not part of the display) and with the reviewer's mismatch judgment for any image that was displayed.

| Outcome | Top ASR hypothesis | Participant's final choice | How it enters the analysis |
|---|---|---|---|
| Useful repair | Wrong | Correct | Primary benefit; compared across the three displays |
| Confirmed correct | Correct | Correct | Time and unnecessary repair actions (replays, alternative selections) are measured here |
| Missed repair | Wrong | Wrong (kept, or swapped for another wrong candidate) | Residual error; split by whether a pictogram was shown, withheld or unavailable |
| Harmful change | Correct | Wrong | Core safety outcome; a visual-induced error when the displayed pictogram matches the wrong selection |

Recordings will use several speakers and two speaking styles so the system is not tuned to a single voice. Because recognition errors can differ systematically between speaker groups [8], error rates and repair outcomes will be reported per speaker. A fixed set of acoustic distortions will generate realistic but reproducible uncertainty, and no speaker from the main evaluation will be used for selecting repair thresholds. We will manually verify the original transcripts and the intended action of each message. The test set will include cases where the recognizer is uncertain but correct, since unnecessary interventions can themselves introduce mistakes.

Each participant will see an underlying message in only one repair condition. We will compare performance on matched item lists, rotating conditions across listeners. The principal semantic measure will require the participant to recover the correct action and its critical arguments. We will analyze recognition confidence, visual mapping quality and repair choice as distinct explanatory factors. The output will include both average improvements and concrete examples where pictograms changed a correct interpretation into an incorrect one. This error analysis is essential for understanding whether visual support adds value to already informative text and audio

# 4. Discussion

A pictogram cannot correct a recognition mistake unless it helps a person identify the intended concept. The proposal treats visual feedback as an optional repair action, with semantic safety and language-dependent coverage as practical constraints. The first experiment will establish feasibility under controlled noise, not the robustness of a deployed multilingual service in clinical use.

The interface does not address all forms of multilingual speech difficulty. Code-switching, colloquial variants and regional vocabulary may have sparse pictogram coverage, and the noise used in a laboratory is not equivalent to daily conversational conditions. Code-switching is a recognized difficulty for ASR [9] and also complicates lookup, because a candidate phrase may mix languages while review is done per language. The exploratory Portuguese extension could draw on existing spoken-language-understanding data [31] for the intent side, leaving pictogram review as the main new cost. The proposal is also complementary to speech-to-pictogram translation [19], [20]: that line of work serves people who rely on pictograms as a primary medium, whereas the repair policy targets any listener who must verify a transcript.

Moreover, a visual confirmation can demand additional attention from the same person the interface is meant to support: text, audio and an image give three sources to integrate for one decision, and cognitive load theory warns that material duplicating information already available can add extraneous load without benefit [35]. We will therefore report time and unnecessary repair actions alongside accuracy, and treat a pictogram that adds seconds without changing outcomes as a negative result. For that reason, optionality and selective intervention are central to the proposed design.